\documentclass[12pt]{article}

\usepackage[nonatbib]{latex/neurips_2019}

\usepackage[utf8]{inputenc} 
\usepackage[T1]{fontenc}    
\usepackage{url}            
\usepackage{booktabs}       
\usepackage{amsfonts}       
\usepackage{nicefrac}       
\usepackage{microtype}
\usepackage{graphicx}
\usepackage[style=numeric,backend=biber]{biblatex}
\bibliography{latex/bibliography}

\title{Shaping Political Discourse using multi-source News Summarization}

\begin{document}

\maketitle

  

\section{Introduction}

Multi-document summarization is an automatic procedure aimed at extraction of information from multiple texts written about the same topic. The resulting summary allows individual users to quickly familiarize themselves with information contained in a large cluster of documents. In such a way, multi-document summarization systems are complementing the news aggregators performing the next step down the road of coping with information overload.

Multi-document summarization creates information reports that are both concise and comprehensive. With different opinions being put together \& outlined, every topic is described from multiple perspectives within a single document. While the goal of a brief summary is to simplify information search and cut the time by pointing to the most relevant source documents, comprehensive multi-document summary should itself contain the required information, hence limiting the need for accessing original files to cases when refinement is required. Automatic summaries present information extracted from multiple sources algorithmically, without any editorial touch or subjective human intervention, thus making it completely unbiased. The multi-document summarization task is more complex than summarizing a single document, even a long one. The difficulty arises from thematic diversity within a large set of documents. A good summarization technology aims to combine the main themes with completeness, readability, and concision.  An ideal multi-document summarization system not only shortens the source texts, but also presents information organized around the key aspects to represent diverse views. Success produces an overview of a given topic.

In an era of increasing political polarization of news agencies, as exemplified by the nature of discourse in the US and India, two of the largest democracies in the world, it has become imperative for the average citizen to be aware of both the sides of the coin before forming an opinion. There exist few news outlets, if at all, that consciously attempt to gather sources from both sides of the spectrum while reporting their stories. Even the ones that do, form the narrative in a manner that is conducive towards directing public discourse in the direction that supports their bias.

In this project, we have designed a machine learning model that takes multiple news documents pertaining to a specific topic as input and generates a concise summary of the topic as the output. The idea is that if the input contains diverse news sources, the output summary would be an amalgamation of most of the arguments and supporting evidence that occur in the summary, thus being inclusive of all the polar viewpoints.

We design an unsupervised model for this problem. Typical unsupervised automatic summarization models are not explicitly designed to capture a diversity of view points. Our model takes a different approach, since it first identifies the relevant parts of the source text that correspond to the various aspects of the topic being reported, and then summarizes each of those aspects individually. Thus, even if 90\% of the news sources pertaining to a topic lean one way, the summary would avoid that bias by sampling its input equally from all the different aspects.


\section{Related Work}
There has been a lot of recent research interest in multi-document summarization, primarily owing to the success of deep learning in the single document case. In \cite{ma2016unsupervised} the authors propose a document-level reconstruction framework named DocRebuild, which reconstructs the documents with summary sentences through a neural document model and selects summary sentences to minimize the reconstruction error. They also apply two strategies, sentence filtering and beamsearch, to improve the performance of their method. 

 \cite{yasunaga2017graph} proposes a neural multi-document summarization system that incorporates sentence relation graphs. They employ a Graph Convolutional Network (GCN) on the relation graphs, with sentence embeddings obtained from Recurrent Neural Networks as input node features. Through multiple layer-wise propagation, the GCN generates high-level hidden sentence features for salience estimation. The authors then use a greedy heuristic to extract salient sentences that avoid redundancy.
 
  More recently,  \cite{liu2018generating} uses a transformer based model to generate summaries, drawing from the recent success of transformers in natural language understanding and generation tasks.  The authors use extractive summarization to coarsely identify salient information and a neural abstractive model to generate the article. For the abstractive model, they introduce a decoder-only architecture that can scalably attend to very long sequences, much longer than typical encoder- decoder architectures used in sequence transduction. They show that this model can generate fluent, coherent multi-sentence paragraphs and even whole Wikipedia articles.  \cite{lebanoff2018adapting} extends methods from single document summarization to multiple documents. Even more recently, \cite{fan2019using} converts the multiple source documents into a knowledge graph before feeding it into a summarization module. 

Argument mining \cite{10.1145/3308560.3316599, jindal2023classification} is an area within natural language understanding that focuses on mining arguments from opinionated text. It is an advanced form of human language understanding by the machine. When sufficient explicit discourse markers are present in the language utterances, the argumentation can be interpreted by the machine with an acceptable degree of accuracy. However, in many real settings, the mining task is difficult due to the lack or ambiguity of the discourse markers, and the fact that a substantial amount of knowledge needed for the correct recognition of the argumentation, its composing elements and their relationships is not explicitly present in the text \cite{singh2019one}, but makes up the background knowledge that humans possess when interpreting language.  Despite its usefulness for this task, most current approaches are designed for use only with specific text types and fall short when applied to heterogeneous texts. In \cite{stab2018cross} the authors propose a new sentential annotation scheme that is reliably applicable by crowd workers to arbitrary Web texts. The authors annotations for over 25,000 instances covering eight controversial topics and show that integrating topic information into bidirectional long short-term memory networks outperforms vanilla BiLSTMs by more than 3 percentage points in F1 in two- and three-label cross-topic settings \cite{rastogiexploring}. 
\cite{stab2017parsing} summarized the state of the field and tried to better organize the various approaches that were being taken into a cohesive structure. Recently,  \cite{reimers2019classification} fine-tuned BERT for classifying arguments for a given topic into for, against or neither and also for measuring similarity between argumentative statements.

\section{Methodology}

In our project, we have have framed the problem of multi-source news summarization as: the system takes a set of news documents pertaining to one topic of interest as input and produces a concise summary of the topic based on the source documents as output. With a focus on being able to generate diverse summaries, we have experimented with two methods for doing this: Unsupervised Pipeline Model and Supervised End-to-End Model. The first approach (Unsupervised Pipeline Model) is a hard problem and we divide it into multiple stages namely - Selection, Clustering and Summarization. The input to the selection step is the set of documents and the output of each of the intermediate steps serves as the input to the next step (it's a sequential model). In the second approach (Supervised End-to-End Model), we use a tweaked version of the BertSum model with a modified loss function. The new loss function incorporates a diversity loss into the original BertSum loss function to capture dissimilar viewpoints in the generated summary \cite{singh2018footwear}. The Supervised End-to End Model takes the ground truth summaries as document labels and optimizes the modified loss function between the generated summaries and the ground truth summaries during training. In the following sections, we further explain the implementation details of the two approaches in detail as well as the challenges.

\subsection{Challenges}

Summarizing news in an unbiased fashion using a model operating on multiple sources poses a few interesting challenges over a models operating on single source:

\begin{itemize}
    \item Multi-document summarization is typically modeled as an extension of the single document version. Concatenating many documents into one produces an input size that is typically too large even for modern transformer based models to handle. Thus, information across documents needs to be condensed, and some recent approaches attempt to do that \cite{fan2019using}. We propose \textbf{selecting a subset of sentences} from the source documents in order to handle this problem. 
    \item The degree of redundancy in information contained within a group of topically-related articles is much higher than the degree of redundancy within an article, as each article is apt to describe the main point as well as necessary shared background. Hence anti-redundancy methods are more crucial in multi-document summarization approaches.
    \item  A group of articles may contain a temporal dimension, typical in a stream of news reports about an unfolding event. Here later information may override earlier more tentative or incomplete accounts. Further, the co-reference problem in summarization presents even greater challenges for multi-document than for single-document summarization.
    \item The model needs to be \textbf{robust to imbalance} in the coverage of the various aspects in the set of source documents pertaining to a topic. 
    \item \textbf{Summarizing} an entire aspect of a topic that has been handled in different ways by different sources is a challenging problem.
    \item The final piece in the puzzle is combining summaries of the various aspects pertaining to a topic and the arguments therein. 
\end{itemize}
We design two types of models which try to solve the above challenges as explained in sections \ref{supervised} and \ref{unsupervised}

\subsection{Unsupervised Pipeline Model} \label{unsupervised}
We design a model that handles the multi-document summarization problem in four stages - a) selection, b) clustering, c) summarization, and d) combination. Each stage corresponds to one of the problems above, and is handled by a separate algorithm: \\

\begin{itemize}
    \item \textbf{Selection}: Select a subset of the original set of sentences from the multiple documents that are most relevant to summary generation. We use \cite{reimers2019classification} which uses a BERT-based model for classifying input sentences as arguments or non-arguments. During summarization, we use either just the arguments, or a combination of argumentative and non-argumentative sentences in a pre-specified ratio and tabulate the results obtained in Section \ref{supervised_results}
   
   \item \textbf{Clustering}: We use K-Means++ for unsupervised clustering on sentence embeddings obtained using BERT. As an alternative, we compute pairwise similarities between argumentative sentences using the argument-similarity BERT model in \cite{reimers2019classification} and use agglomerative hierarchical clustering to cluster the sentences. 
   
   \item \textbf{Summarization}: For K-Means clusters, we choose one sentence per cluster that lies closest to the centroid for that cluster. For agglomerative clusters, we use the sentence with maximum cumulative within-cluster similarity score. 
   
   \item \textbf{Combination}: Combination has been handled by concatenating the summaries obtained from the previous step.
\end{itemize}

\begin{figure}
    \centering
    \includegraphics[width=\textwidth]{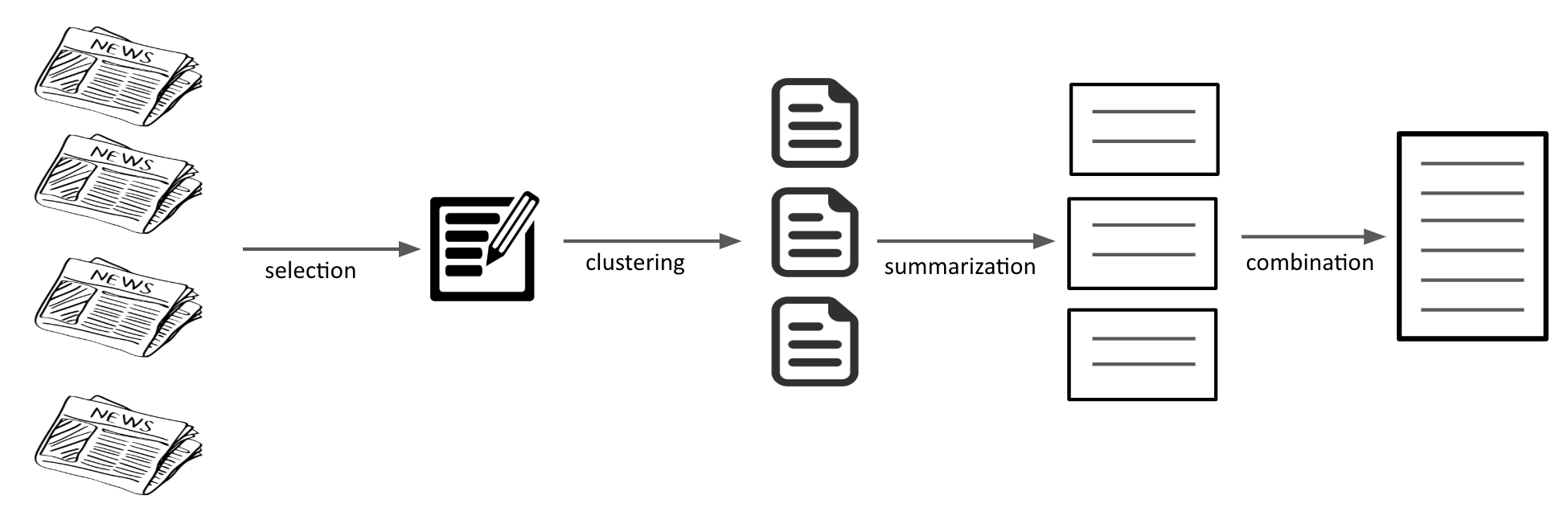}
    \caption{Model Schematic: The multi-document summarization model attempts to balance summary quality with argument diversity for hot news topics. It is composed of four stages, each of which is governed by a separate algorithm: a) selection of sentences from source news articles, b) clustering of sentences based on their theme, c) summarization of individual clusters, d) combination of summaries to form a single summary for the topic.}
    \label{fig:model}
\end{figure}

The above steps have been pictorially represented in Fig. \ref{fig:model}.



\subsection{Supervised End-To-End Model} \label{supervised}

\begin{figure}
    \centering
    \includegraphics[width=\textwidth]{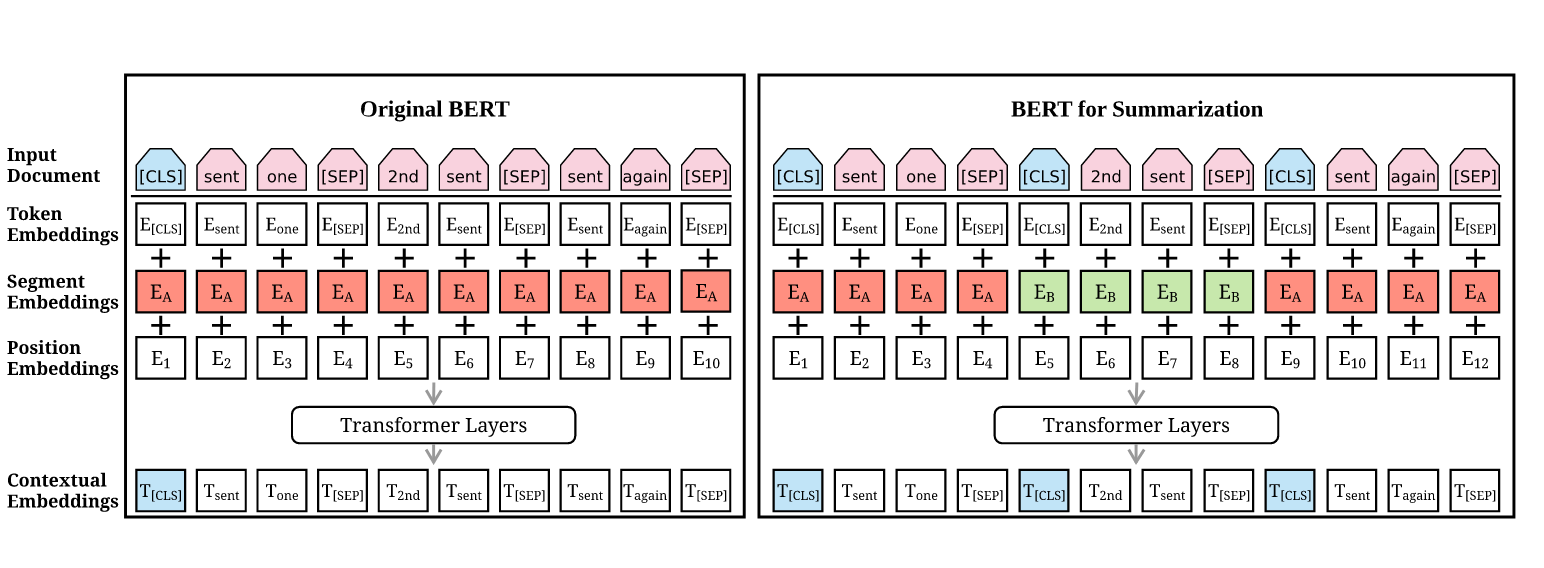}
    \caption{BertSum model diagram as in \cite{liu2019text}. }
    \label{fig:bertsum}
\end{figure}

Most supervised methods for extractive multi-document summarization rely on concatenating the multiple documents and training on the cross-entropy loss between the selected sentences and the gold-standard summary sentences. We tweak BertSum, a model developed in \cite{liu2019text} to better suit our objective of making our summaries as inclusive of diverse opinions as possible. 

BertSum uses a BERT-based architecture to encode each sentence in an input document. Further, a transformer model is used on top of the sentence encodings themselves (treating the sentences in a document like words in a sentence) to get top-level sentence embeddings. These top level embeddings are then used for binary prediction for inclusion/exclusion of a sentence in the summary. 

We modify the binary cross-entropy loss function to include a term to account for the diversity of the sentences being picked. We define the "diversity penalty" for a document as the weighted sum of similarities between all pairs of sentences within the document. The weight is determined by the product of the probabilities assigned to the two sentences by the BertSum model for inclusion in the extractive summary. This encourages the model to select sentences in the summary that are diverse (and consequently contain different sides of various arguments).

\begin{equation}
    Loss = \sum_{i=1}^{N} BCE(v_{i}, \hat{v}_{i}) + \sum_{i=1}^{N} \sum_{j=1}^{N} \hat{v}_{i} \hat{v}_{j} sim(i, j)
\end{equation}{}
Here, $v_{i}$ corresponds to the ground truth value for the $i^{th}$ sentence (1 for inclusion in the summary, 0 otherwise). $\hat{v}_{i}$ corresponds to the predicted probability by the BertSum model for inclusion in the summary, and BCE corresponds to the binary cross entropy loss. 

\begin{equation}
    sim(i, j) = h_{i}^{L} * h_{j}^{L}
\end{equation}{}

where $h_{i}^{L}$ refers to the sentence embedding of the $i^{th}$ sentence after having gone through the BERT encoder and the inter-sentence transformer and the multiplication operator refers to the dot product. Fig. \ref{fig:bertsum} summarizes the BertSum model used in the Supervised End-to-End model approach.

\section{Experiments and Analysis}
\subsection{Dataset}

We have used two datasets to implement, train and evaluate the pipeline model and the end-to-end model. For the unsupervised pipeline model where we have optimized various stages of the pipeline separately, namely, selection, clustering and summarization, we have used the standard multi-document summarization dataset - the Document Understanding Conference (DUC) dataset. We have used DUC-2001-03 datasets for training and DUC-2004 dataset for evaluation. 

Similarly, for the supervised end-to-end model, we have used the DeepMind Q\&A Dataset which consists of CNN and Daily Mail datasets which have been created from news articles for Q\&A research. The CNN dataset contains the documents and accompanying questions from the news articles of CNN. There are approximately 90k documents and 380K questions. The Daily Mail dataset contains the documents and accompanying questions from the news articles of Daily Mail. There are approximately 197K documents and 879K questions. We have trained the end-to-end model on the CNN and Daily Mail datasets and have evaluated it on the DUC-2004 dataset as well as a manually curated dataset of 5 controversial topics consisting of 5 polarised news documents each. The details of the manually curated dataset are explained in Section \ref{manual}

\subsubsection{Manually Created Dataset} \label{manual}

To evaluate the supervised end-to-end model on a new dataset, we manually create a dataset consiting of five controversial and recent topics and curate five short news articles for each of these topics. The five news articles were sourced from different news article agencies like New York Times, BBC, CNBC, The Guardian etc. and consist of polarized viewpoints about the main topic. The topics selected were:  Citizenship Amendment Bill in India, UK elections, US government shutdown, Forbes list of most Powerful Women and Trump impeachment. The intuition behind sourcing the news articles from different news agencies was to capture polarized viewpoints each news agency might have for the topic in question and analyze the diversity of summaries our two approaches were generating. We have implemented the Unsupervised Pipeline model on one of the topics from this dataset and shown the result in \ref{example}. 

\subsection{Baseline and Oracle}
As mentioned in the project proposal, for the baseline, we had generated sentence embeddings by using Bert and had clustered using KMeans++ algorithm \cite{kumari2017parallelization} after which we chose the sentence having its embedding closest to the cluster centroid. We also considered the oracle to be a human comprehensively and accurately writing the summaries of various documents. Our baseline did not give extra considerion to the arguments present in the various articles. 

\subsection{Unsupervised Pipeline Model} \label{supervised_results}
We train the Unsupervised Pipeline Model on the DUC-2001-03 datasets and then test the approach on five different topics from the DUC-2004 dataset. Each topic has about 10 articles and has 4 summaries written by humans. We randomly select one of the four summaries as our ground-truth summary.

For the first step of the Unsupervised Pipeline Model, aka, selection, we used UKPLabs BERT-argument-classification-and-clustering \cite{bert-argument-classification} to select the arguments from the given documents. In the and for extractive summarization, we used bert-extractive-summarizer  \cite{bert-extractive-summarizer} which takes the text to be summarized along with the ratio of sentences to be present in the summary. Firther, for the clustering step, we used K-Means++ for unsupervised clustering on sentence embeddings obtained using BERT. As an alternative, we compute pairwise similarities between argumentative sentences using the argument-similarity BERT model in [1] and use agglomerative hierarchical clustering to cluster these sentences. During the sumamrization phase, for KMeans++ clustering, we choose one sentence per cluster that lies closest to the centroid for that cluste whereas  for agglomerative clusters, we use the sentence with maximum cumulative within-cluster similarity score. 
We further run 3 different types of experiments - full, only arguments and mix to evaluate different combinations and percentage weightage of the arguments in the final summaries. \\ \\In the full experiment, we combine the sentences from all the ten articles and run extractive summarization to get the final summary. As expected, here, we are not doing any selection or clustering and are directly implementing the summarization step (step 3). In the second experiment which is only arguments, we combine all the ten articles but only select the arguments selected by the argument classifier. We then run extractive summarization on the identified arguments to get to the final summary. Hence, here we are just implementing step 1 (selection) and step 3 (summarization) and skipping step 2 (clustering). In the mix experiment, we combine all the ten articles and make two sets - the sentences classified by the argument classifier as 'arguments' and the sentences classified by the argument classifier as 'non-arguments'. We summarize each of these sets separately and later concatenate the two summaries to get the final summary. We choose different levels of mixing between argument and non-argument sentences by constraining the length of the summaries to different values. We choose lengths for non-arguments set summary to be 220, 330, 440 characters, (which we refer to hereafter by 33.3\% mix, 50\% mix and 66.7 \% mix respectively)  and correspondingly for the arguments set summary to be 440, 330, 220 characters.\\ \\ Restricting the summary to 660 characters (or in case of mix to 220/330/440 characters) was a challenging task and we used heuristics. The extractive summarization works on the ratio of sentences desired to be present in the final summary and thus as the length of the sentences varied a lot, we constrained the problem to finding sentences of lengths less than 200 characters. We relaxed this constraint for summarizing arguments when there were not many arguments present in the set of articles (when the articles were more a description of an event). We tabulate the different models, precision and recalll scores in Table \ref{tab-unsupervised}.

\begin{table}[h]
  \caption{Rouge-1 scores for Unsupervised Pipeline Model}
  \label{tab-unsupervised}
  \centering
  \begin{tabular}{lll}
    \toprule

    Model     & Recall & Precision \\
    \midrule
 All sentences + KMeans++ & 0.2377 & 0.2253 \\ 
 Only Arguments + KMeans++ & 0.1591 & \textbf{0.2354}  \\
 ArgsKM 66.7\% & 0.2133 & 0.2219  \\
   ArgsKM 50\% & 0.2262 & 0.2145  \\
 ArgsKM 33.3\% & \textbf{0.2425} & 0.2038  \\ 
 Only Arguments + Agglomerative & 0.195 & 0.177 \\ 

    \bottomrule
  \end{tabular}
\end{table}

\subsubsection{Error Analysis and Discussion}
We observe that the precision score for the Only Arguments + KMeans++ is the maximum whereas the mix experiment with 33.3\% arguments and 66.7\% no arguments gave the highest recall. The Only Arguments + KMeans++ experiment has the highest precision as intuitively any summary is a brief collection of the main arguments presented in an article plus some non-argument sentences. Hence, through this experiment, we are actually just identifying and selecting the main arguments in the article and clubbing them together to generate the summary. This means, we are not including any non-argument sentences and our summary only contains factual/strong argument holding sentences. Hence, there is a high likelihood of these generated summaries to be very close to the gold-standard summaries and all the arguments selected are definitely present in the gold-standard summary. The articles in the DUC dataset have fewer arguments and are more factual. Thus the arguments do not comprehensively cover the summary of the articles. Hence only arguments experiment has a significantly low recall score. Similarly, for the Only Arguments + KMeans++ experiment, the recall is the maximum because there is a high overlap between the gold-standard summaries (which have a mix of arguments and non-argument sentences) and our generated summaries (from the miox experiment). Intuitively, this emphasizes that the percentage of argument and non-argument senteces in the gold standard summaries is close to 33.3\%. We also see that as the percentage of non-arguments summary length increases, the recall increases but the precision decreases. 
\subsection{Supervised End-to-End Model}
We start with a BertSum model pre-trained on the CNN/Daily Mail dataset (\texttt{CNN/DM BertExt} at \cite{bertsum}). This model is  fine-tuned on the same dataset using the modified loss with the diversity penalty for 2000 steps. This is the end-to-end model that we use here. \\
We test on 5 different topics from DUC-2004. Each topic has about ten articles and has four summaries written by humans. We randomly select one of the four summaries as our ground-truth summary.
The end-to-end model was pretrained to work on documents with number of tokens less than 510. As the concatenation of the ten articles would have much larger number of tokens, we randomly select five of those articles and then concatenated approximately the first 100 words from each selected article. We do this as we expect that the first part of an article would contain more number of arguments and would contribute more to the summary as compared to later parts of the article. \\We preprocessed the resulting documents as done in \cite{bertsum} where we tokenized the document, converted it to json and then generated pytorch files which we were finally fed into the model. We ran both the original model with the original loss function  and modified model with the diversity loss function incorporated and provide the Rouge-1 precision and recall scores in Table \ref{tab:supervised}.

\begin{table}[h]
  \caption{Rouge-1 scores for end-to-end model}
  \label{tab:supervised}
  \centering
  \begin{tabular}{lll}
    \toprule

    Model     & Recall & Precision \\
    \midrule
    BertSum & 0.240 & 0.289 \\
    BertSum with Diversity loss & 0.231 & 0.263  \\

    \bottomrule
  \end{tabular}
\end{table}

\subsubsection{Error Analysis and Discussion}
We notice that both the models - original BertSum model and BertSum with Diversity loss model have higher Rouge-1 precision and recall scores than most of the unsupervised pipeline models. This might be because the pretrained BertSum model is highly tuned for single document extractive summarization which would also work on the concatenation of multiple documents. \\
Recall and precision scores decrease on introducing the diversity loss. This might be due to the DUC dataset containing more factual sentences instead of arguments and so the gold summaries not incorporating diversity. We also observe that the recall score is noticeably smaller than the precision value for both models. This is explained by the fact that we took only the first part of half of the articles in each topic which would result in lower recall. 


\section{Example} \label{example}

To better explain the selection, clustering and summarization stages, we run our pipeline model on one of the topics of our manually created dataset which consists of five recent news topics. We chose the topic 'Citizenship Ammendment Bill in India' which contains 5 short news articles. \\
One of the articles is "Tens of thousands of protesters rioted in three states across India’s northeast, some defying a government curfew and military deployment to demonstrate against the passage of the contentious Citizenship Amendment Bill, which will grant citizenship to thousands of migrants on religious grounds .... Critics fear the bill will be used to harass Indian Muslims by forcing them to pass a citizenship test and prove their family’s lineage in the country, while giving a blanket pass to people of most other religions.But government officials say the bill is a humanitarian effort to provide shelter to religiously persecuted minorities. The bill is expected to be signed into law in the coming days."

We run the 5 short news articles through our pipeline model. The total number of sentences is 57. The number of arguments selected by the model is 10. Examples of the arguments selected are - "Central to the idea was that your religious identity would be irrelevant to your belonging, and it's that which is being turned on its head.", "Critics fear the bill will be used to harass Indian Muslims by forcing them to pass a citizenship test and prove their family’s lineage in the country, while giving a blanket pass to people of most other religions." 

We run pariwise similarities between all the 57 sentences and use agglomerative clustering to cluster the sentences into 5 clusters. An example of cluster in this topic can be one having sentences related to protests and demonstrations. Some of the sentences included by our model in this cluster are - "By Thursday night, the government had shut down the internet, deployed hundreds of troops, imposed a curfew in Assam state and banned groups of more than four people from assembling in neighboring Meghalaya state.",
"The police shot and killed two protesters in Assam whom they accused of defying the curfew, and arrested dozens of others there, The Associated Press reported.",
"Protesters are angry that the bill will grant citizenship to thousands of Hindu, Christian, Jain, Buddhist and Sikh migrants from some neighboring countries where New Delhi says they are religiously persecuted."\\ \\
Finally we present a summarization in which we use no selection, agglomerative clustering and extractive summarization. The summary obtained is "Mander said the very nature of the Indian constitution is that it is based on secular values. " Demonstrators say this will flood their hometowns with unwanted foreigners. Critics fear the bill will be used to harass Indian Muslims by forcing them to pass a citizenship test and prove their family’s lineage in the country, while giving a blanket pass to people of most other religions. But government officials say the bill is a humanitarian effort to provide shelter to religiously persecuted minorities. The bill is expected to be signed into law in the coming days." \\ \\ We can observe that this short summary contains both the arguments supporting and opposing the Citizenship Ammendment bill and can help reduce bias towards any one point of view.


\section{Social Impact}
This project would help users to get a summary of any news topic containing all the various points of views and the different arguments given in a set of articles. Reading an article emphasizing only one side of an argument can increase the inherent bias that a person might have. By emphasizing on diversity of arguments, this project would help users to reduce their inherent bias \cite{10.1007/978-981-10-8639-7_25}. This project also tries to place the same weight on similar arguments on a topic which are present in several articles and on another argument in a single article on the same topic. This would help reduce people's fear of news articles maliciously discarding a point of view.

\section{Code}
The repository containing the code and sample of the data can be found at https://github.com/c-rajan/Multi-source-NewsSummarization

\section{Conclusion}

The main motivation behind this project was to develop a multi-document summarization framework for news articles. We have ideated, implemented and experimented two different approaches for the same: Unsupervised Pipeline Model and Supervised End-to-End Model. The Unsupervised Pipeline Model was split into various step: selection, clustering, summarization and combination. We implemented each of these steps and tabulated the results for the same. In the second approach - Supervised Pipeline Model, we implemented an end-to-end model which leverages the BerSum model and generates extractive summaries of the given documents. We tweak the BertSum model by adding the diversity loss function in the original BertSum model loss function and tabulate the precision and recall scores for both these approaches. We conclude that the second approach (Supervised End-to-End Model) has a higher precision and recall score as compared to the Unsupervised Pipeline Model, and more specifically, the original BertSum model without the diversity loss function incorporated has a higher precision score as compared to the BertSum model with diversity loss. This means that, the gold-standard summaries of the DUC 2004 dataset have less diversity in their content.

\printbibliography
\end{document}